\documentclass[sigconf,nonacm]{acmart}
\usepackage{latexsym}
\usepackage{amsmath}
\usepackage{graphicx}
\usepackage{booktabs}
\usepackage{multicol}
\usepackage{multirow}
\usepackage{amsmath}
\usepackage{amsfonts} 
\usepackage{subfigure}
\usepackage{marvosym}
\usepackage{algorithm}
\usepackage{algorithmic}
\usepackage{subcaption}
\usepackage{tabularx} 
\usepackage{array}    
\usepackage{geometry} 
\usepackage{diagbox} 
\AtBeginDocument{%
  }

\begin{document}

\title{RLAP: A Reinforcement Learning Enhanced Adaptive Planning Framework for Multi-step NLP Task Solving}


\author{Zepeng Ding}
\affiliation{%
  \institution{KW, SDS, Fudan University}
  \city{Yangpu District}
  \state{Shanghai}
  \country{China}}
\email{dingzepeng@fudan.edu.cn}

\author{Dixuan Wang}
\affiliation{%
  \institution{KW, SDS, Fudan University}
  \city{Yangpu District}
  \state{Shanghai}
  \country{China}}
\email{dxwang23@m.fudan.edu.cn}

\author{Ziqin Luo}
\affiliation{%
  \institution{KW, SDS, Fudan University}
  \city{Yangpu District}
  \state{Shanghai}
  \country{China}}
\email{zqluo22@m.fudan.edu.cn}

\author{Guochao Jiang}
\affiliation{%
  \institution{KW, SDS, Fudan University}
  \city{Yangpu District}
  \state{Shanghai}
  \country{China}}
\email{100621008@qq.com}

\author{Deqing Yang}
\affiliation{%
  \institution{KW, SDS, Fudan University}
  \city{Yangpu District}
  \state{Shanghai}
  \country{China}}
\email{yangdeqing@fudan.edu.cn}

\author{Jiaqing Liang\textsuperscript{\Letter}}
\affiliation{%
  \institution{KW, SDS, Fudan University}
  \city{Yangpu District}
  \state{Shanghai}
  \country{China}}
\email{liangjiaqing@fudan.edu.cn}

\renewcommand{\shortauthors}{Ding et al.}

\begin{abstract}
Multi-step planning has been widely employed to enhance the performance of large language models (LLMs) on downstream natural language processing (NLP) tasks, which decomposes the original task into multiple subtasks and guide LLMs to solve them sequentially without additional training. 
When addressing task instances, existing methods either preset the order of steps or attempt multiple paths at each step. 
However, these methods overlook instances' linguistic features and rely on the intrinsic planning capabilities of LLMs to evaluate intermediate feedback and then select subtasks, resulting in suboptimal outcomes.
To better solve multi-step NLP tasks with LLMs, in this paper we propose a \textbf{R}einforcement \textbf{L}earning enhanced \textbf{A}daptive \textbf{P}lanning framework (RLAP). 
In our framework, we model an NLP task as a Markov decision process (MDP) and employ an LLM directly into the environment. In particular, a lightweight \emph{Actor model} is trained to estimate Q-values for natural language sequences consisting of states and actions through reinforcement learning. 
Therefore, during sequential planning, the linguistic features of each sequence in the MDP can be taken into account, and the Actor model interacts with the LLM to determine the optimal order of subtasks for each task instance. We apply RLAP on three different types of NLP tasks and conduct extensive experiments on multiple datasets to verify RLAP's effectiveness and robustness.
\end{abstract}

\ccsdesc[500]{Computing methodologies~Natural language processing}

\keywords{Reinforcement Learning, Large Language Model, Natural Language Processing}


\maketitle

\section{Introduction}
At present, large language models (LLMs) are widely used in many basic natural language processing (NLP) tasks, such as information extraction (IE), common-sense question answering (QA), text generation, etc. However, when solving the problems in practical scenarios, LLMs frequently encounter the contexts with diverse grammatical structures and complex semantics, as well as multiple types of personalized questions, such as complex domain-specific information extraction and multiple-choice QA. It is difficult to solve these tasks in one step by directly employing LLMs with prompts. For example, LLMs exhibit a low recall rate when handling complex information extraction~\cite{ding2024improving} and generate hallucinatory responses during the reading comprehension process~\cite{yao2023llm}.

\begin{figure}[!htb]
\centering
\subfigure[]{
\includegraphics[width=0.95\linewidth]{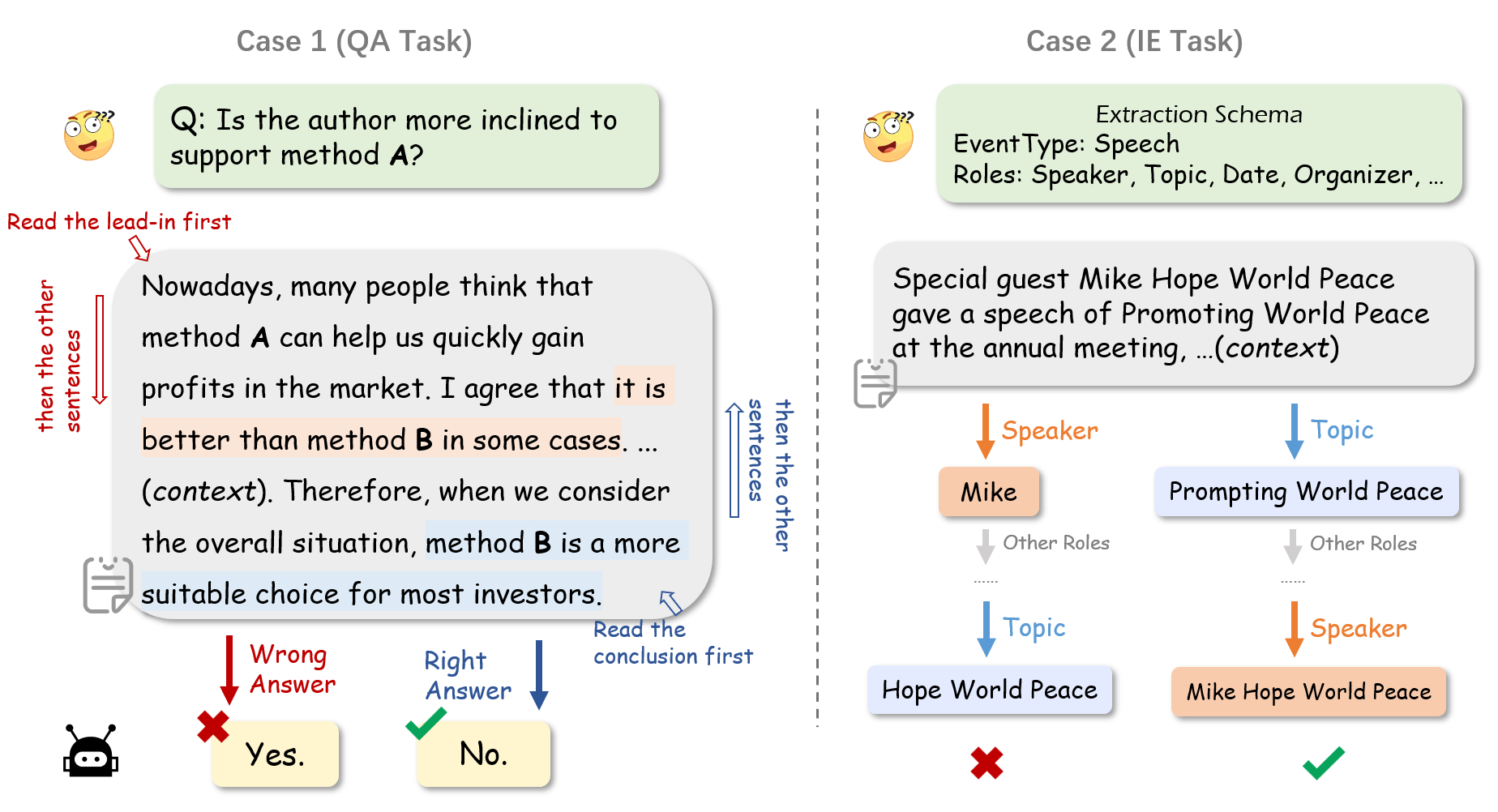}
\label{fig:intro_1}
} \\
\subfigure[]{
\includegraphics[width=0.95\linewidth]{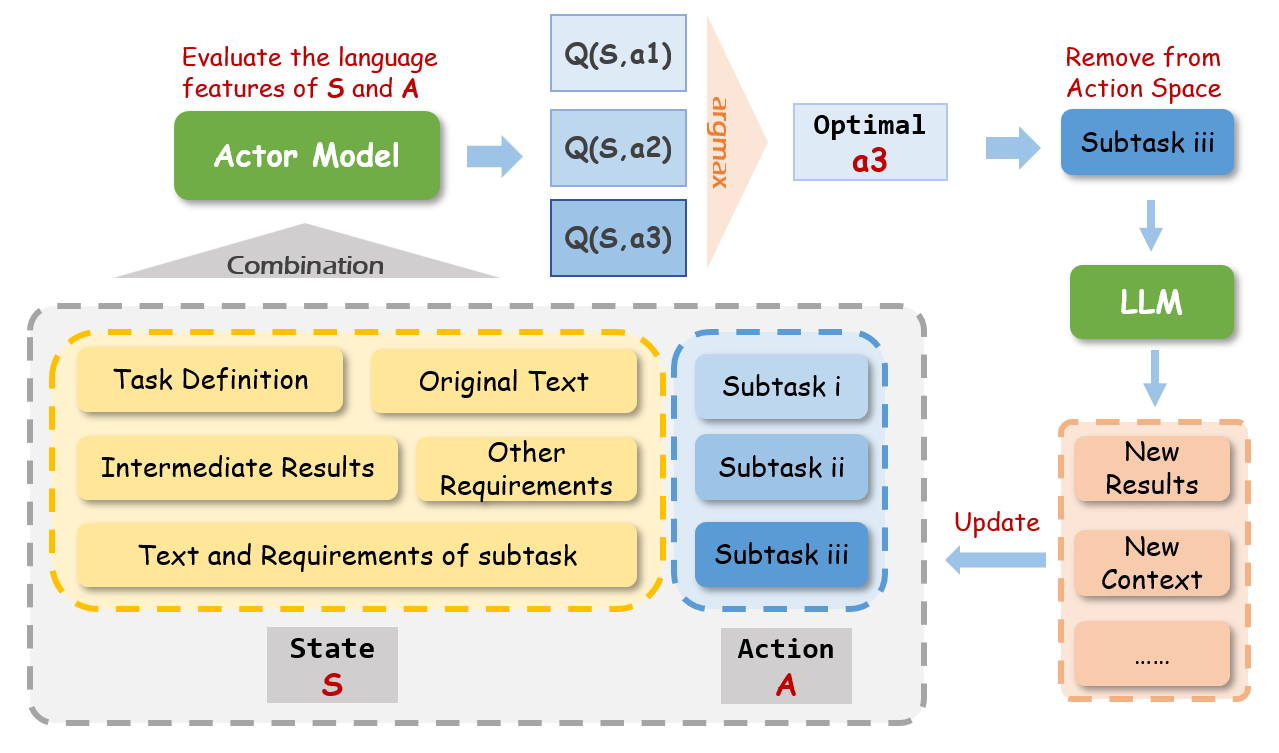}
\label{fig:intro_2}
}
\vspace{-0.3cm}
\caption{(a) For the same task instance, different action orders will lead to different results. (b) Overview of the proposed RLAP.}
\label{fig:intro}
\vspace{-0.2cm}
\end{figure}

To enable LLMs to achieve satisfactory performance on specific NLP tasks, prevalent methods focus on fine-tuning or preference alignment on additional labeled task data~\cite{zhang2024scaling, rafailov2024direct}. However, these methods cannot be applied to closed-source LLMs like ChatGPT and have notable drawbacks.
First, fine-tuning LLMs for specific tasks using labeled data requires substantial computing resources and time consumption, which is not feasible for most users.
Second, different types of NLP tasks typically require different fine-tuning procedures. 
Third, the performance enhancement achieved through supervised fine-tuning (SFT) is constrained by the tasks' complexity, the quality of data, and the training framework. 
Training on specific task data leads to degradation in general knowledge of LLMs~\cite{luo2023empirical}.

Some previous work has explored breaking down one task into multiple subtasks and gradually guiding the LLM to solve the problem without fine-tuning, such as ChatIE~\cite{wei2023zero}. Further research introduced the evaluation of intermediate results and interactive feedback when solving problems in multiple steps, such as ReAct~\cite{yao2023react}, AutoScraper~\cite{huang2024autocrawler} and ToT~\cite{yao2024tree}.
For different task instances, these methods predetermine the order of subtasks before solving them (such as reading from the first sentence to the end when processing long text), or make multiple attempts at each step (such as trying all possible sub-HTML reading orders or query orders in webpage QA). 
Notably, these methods overlook the influence of task/subtask instances' \textit{linguistic features} (such as \textit{semantics} and \textit{syntactic structures}) on results.
In fact, LLMs' performance on many NLP tasks is significantly affected by subtask orders, while the optimal order of subtasks depends on the linguistic features and intermediate feedback of task instances~\cite{ding2024adaptive}.
For example, as shown in Fig.~\ref{fig:intro_1}, the reading order can affect the answer to a specific question, and different extraction orders also have an impact on the recognition of complex entities. The linguistic features corresponding to different instances of the same task type are very diverse and difficult to be classified explicitly. 
Therefore, when choosing and solving subtasks, the order should not be completely determined according to the task type, nor should all possible orders be executed exhaustively by LLMs. Instead, we need to take into account the linguistic features of the original text and intermediate feedback to sequentially plan subtask orders.

In addition, these methods rely on the inherent planning ability of LLMs to evaluate intermediate feedback, and then select and solve the next subtask. In each step, an LLM checks the intermediate feedback and then plans and executes the next action. 
Since the result of the previous step affects the selection of next subtask, the LLM's planning and execution processes are coupled. However, the inherent planning ability of LLMs is limited, leading to the accumulation of cross-step errors.
In addition, it is difficult to cope with diverse linguistic features by relying solely on LLMs' inherent planning ability, resulting in poor generalization ability between different task instances.

In this paper, we find that the process of solving subtasks sequentially can be naturally modeled as a Markov decision process (MDP), where the state consists of the current text, task requirements, intermediate results, etc., and the action in each step is selecting the next subtask. Therefore, we propose a \textbf{R}einforcement \textbf{L}earning Enhanced \textbf{A}daptive \textbf{P}lanning Framework for LLMs to solve various NLP tasks, namely \textbf{RLAP}, which uses an LLM (without fine-tuning) as a "task executor", and introduces a separate model to quantitatively assess the states with different linguistic features, and thus helps the LLM plan to complete the MDP.
With the goal of selecting the optimal action (the next subtask to be solved) based on the current state, we further design a lightweight Actor model that estimates Q-values based on sentence-level embeddings and selects the next action precisely. We obtain rewards based on the LLM's feedback on the previous subtask and apply the RL framework to train the Q-value approximation. 
After training, the Actor model can adaptively select the next subtask to be performed based on linguistic features at each step, thereby reducing error accumulation and guiding the LLM to achieve better performance, as shown in Fig.~\ref{fig:intro_2}. For different types of NLP tasks, it is not required to fine-tune the LLM multiple times. Instead, we define the state and action space separately and train an Actor model for each task.

In our experiments of demonstrating how to model the process of solving NLP tasks as MDP, we select three representative NLP tasks, i.e., machine reading comprehension (MRC), IE (including triple extraction, event extraction, etc.) and sentence-level text completion. 
The experimental results reveal that, by introducing RLAP, LLMs' performance on these tasks is significantly improved compared to the fixed-order multi-step solutions, and their performance remains stable when the task instances' complexity increases.


Our major contributions in this paper are summarized as follows:

\begin{itemize}
\item For the NLP tasks that can be modeled as MDPs, we propose a general multi-step solution framework RLAP based on reinforcement learning, which models the linguistic features of the initial and intermediate steps and adaptively guides LLM to plan the next action.

\item We select three practical NLP tasks to confirm that the order of subtasks has an impact on the final results of LLMs, and demonstrate how to define reasonable states and actions based on the types of tasks and subtasks. We also design corresponding Actor models for each task type, which estimate Q-values and select the optimal actions, and are trained according to RLAP.

\item Our extensive experimental results on different task types and different languages demonstrate the effectiveness of our proposed RLAP, which significantly outperforms the baselines on evaluation metrics corresponding to various tasks, and performs stably on complex instances. The ablation study also verifies the necessity of the Actor model.
\end{itemize}

\section{Related Work}
\subsection{Multi-step Planning for LLMs}
Recently, prompting LLMs to do multi-step planning has been applied in many tasks and scenarios, including LLM-based agents~\cite{huang2022inner,ding2023task,xie2024travelplanner}. The most direct approach is to plan a fixed sequence of actions according to the task type. For example, ChatIE~\cite{wei2023zero} transforms IE task into a multi-turn QA task and sets subtask sequences for different extraction schemas; ReAct~\cite{yao2023react} considers reasoning and thinking as independent subtask steps.
Some methods also combine prompt guidance with appropriate data structures and search methods to make multi-step planning more effective, evaluate intermediate feedback, and adjust the search strategy. For instance, AutoScraper~\cite{huang2024autocrawler} leverages the hierarchical structure of HTML and similarity across different web pages to progressively process sub-HTML and synthesize answers; ToT~\cite{yao2024tree} and RAP~\cite{hao-etal-2023-reasoning} use basic tree search and Monte Carlo tree search to search and plan the optimal action, respectively.
As described in Section 1, these methods either specify a fixed order ~\cite{wei2023zero,yao2023react} or make multiple attempts at each step~\cite{yao2024tree,hao-etal-2023-reasoning,huang2024autocrawler}. The evaluation of intermediate feedback and the planning of the next step are highly dependent on the intrinsic capabilities of LLMs. Some works utilize external planners for multi-step planning, such as LLM+P~\cite{liu2023llm+} and LLM-DP~\cite{dagan2023dynamic}, but using these methods requires converting instructions into specific programming languages, and code-based planning is constrained by its narrow domains and the predefined environment~\cite{hao-etal-2023-reasoning}.
 
\subsection{Reinforcement Learning in NLP Tasks}
Deep reinforcement learning (DRL)~\cite{mnih2015human} is mainly applied to intelligent agents that learn to reason in Markov decision processes (MDPs). In recent years, DRL methods have gained increasing attention in the NLP field, including dialog~\cite{li2016deep}, coreference~\cite{yin2018deep}, information extraction~\cite{huang2023adaptive,qin2018robust}, text classification~\cite{wu2018reinforced}, etc. Many NLP tasks can be formulated as MDPs involving incremental decision making~\cite{wang2018deep}.

For large language models (LLMs), RL is often used to help align model output with human preferences. For example, Reinforcement Learning with Human Feedback (RLHF)~\cite{stiennon2020learning} introduces a paradigm for leveraging RL to improve downstream performance on generative language tasks~\cite{kreutzer2018reliability, ziegler2019fine}, achieving great success in ChatGPT~\cite{openai2023gpt4} and Llama2~\cite{touvron2023llama}; RLGF~\cite{chang2023learning} presents a unified framework of incorporating a guiding policy to enhance RL for language generation and outperforms PPO for fine-tuning LLMs. These methods mainly focus on dialogue and language generation and use RL methods to fine-tune LLMs to improve their general performance. 
Some works explore combining LLM and RL algorithms to solve specific NLP tasks, for instance, bilevel-LLM~\cite{liu2024best} incorporates fine-tuning LLM with RL to address the key limitations of traditional RL methods in medical dynamic treatment regime, and \cite{gholamian2024reinforcement} describes the NLP task as an MDP in the prompt to instruct LLM in problem-solving. Moreover, iterative prompting can be characterized as an MDP that captures the interaction between a prompt provider $\pi$ and a language model environment $E$~\cite{yang2023foundation}. Various prompting schemes can be characterized by high-level actions that map input strings to desired output strings using the language model~\cite{press2023measuring,kumar2021reordering,shi2023large}, and these actions can also be combined recursively to implement more complex iterative prompting schemes~\cite{zhou2022least}.

\section{Framework of RLAP}
In this section, we introduce the detailed process of RLAP as well as the structure of each module in RLAP. RLAP can be applied to the NLP tasks that can be decomposed into several subtasks and carried out in multiple steps. 
Initially, we define the states and actions for these NLP tasks in an abstract and unified manner, based on which the MDP is constructed. We also introduce the design and training of the Actor model. Finally, we explain how the LLMs interact with the Actor model to accomplish the task. The application of RLAP to some specific types of NLP tasks is presented in Chapter ~\ref{sec:application}.

\subsection{Preliminaries}

In the context of sequential decision making under uncertainty, the problem is often formalized in terms of a (finite) Markov Decision Process (MDP), which is defined by a tuple $\mathcal{M}:=(S, A, P, R, \gamma)$ consisting of a state space $S$, an action space $A$, a transition probability function $P(s'|s,a): S\times A\times S \to [0,1] $, a reward function $R(s,a): S \times A \rightarrow \mathbb{R}$, and a discount factor $\gamma \in [0, 1]$. The objective in an MDP is to find a policy $\pi: S \to A$ that maximizes the expected cumulative reward over time, typically expressed as the expected return $\mathbb{E}\left[\sum_{t=0}^H \gamma^t R(s_t, a_t)\right]$, where $t$ is the time step and $H$ denotes total steps in an episode (trajectory) $\tau:=\{(s_0,a_0,r_0),...,(s_H,a_H,r_H)\}$.

Unlike methods that fine-tune or rely on LLM for planning, in our framework, we directly embed LLM into the environment without any training or parameter modification. 
Instead, we use a lightweight foundation model with language representation capability (i.e., the Actor model) and train it to learn sequential decision making from interaction experience with the environment, as in traditional DRL. 
Standard reinforcement learning aims to maximize the expected returns of a policy, and we use the value-based methods in RLAP, which are typically more sample efficient than policy-based methods~\cite{gu2016q}. For instance, Q-learning~\cite{watkins1992q}  involves learning the optimal value function $Q^*(s_t,a_t)$ by satisfying a set of Bellman optimality constraints:
\begin{equation}
\label{eq:Qlearn}
	Q^*(s_t,a_t) = r_t+\gamma \mathbb{E}_{s_{t+1}\sim P(s_{t+1}|s_t,a_t)} [\max_{a_t+1}Q^*(s_{t+1},a_{t+1})],
\end{equation}
and an optimal policy can be obtained: $\pi^*(\cdot|s_t)=\arg\max_a Q^*(s_t,a)$.

In RLAP, we define a series of subtasks $A=(a_1,..,a_k)$ according to the type of NLP task, and these subtasks constitute the action space. Within a complete episode, the LLM sequentially addresses each subtask and ultimately arrives at the final answer, so the total steps $H$ is equal to the number of subtasks $k$.
Each action can be executed independently by LLM, but the result of the previous step will affect the current action selection and then affect LLM's output.
The objective of the LLM at each step is to accurately address the subtask within the current state based on the previous results and the task requirements, while the Actor model is required to select the most appropriate subtask from the remaining actions for the next step. 

Our training objective is that, for a given task instance, the Actor model is able to plan an optimal sequence of actions $A^* = (a_1^*, ..., a_k^*)$, which is a permutation of $A$, such that the LLM attains the correct answer after executing all the subtasks in accordance with $A^*$. The overall goal of the framework is to enhance the performance metrics of various task instances within a certain type of NLP task.

\subsection{Environment Construction}

Sequentially solving tasks according to RLAP is the process of the Actor model as an agent interacting with the environment embedded with the LLM. 
We first introduce the composition of the MDP environment in RLAP, and for different task types, we need to define different states, action spaces, and reward functions, which are illustrated in detail in Chapter ~\ref{sec:application}.
\paragraph*{State}
Each state is defined as a \textbf{dictionary}, which contains task definition, original text, intermediate results, other requirements such as format and content, and the subtask description, as shown in Fig.~\ref{fig:intro_2}.
At each step, we construct the prompt for the LLM based on the state, and after the LLM completes the subtask, we calculate the reward, concatenate or modify the intermediate results, and construct the next state.
Compared with methods that utilize text to store interaction information, the state in RLAP does not necessitate the storage of historical prompts and interaction records. Beyond global information such as the original text and task definition, it solely requires the storage of intermediate results and the requirements of the current subtask. Consequently, the length of the prompt does not increase appreciably as the number of interaction steps grows.
If the final result is a concatenation of the results from each step, the length of the prompt only increases by the number of tokens concatenated in each step. This renders the inferring process of the LLM more cost-effective and efficient.

\paragraph*{Action}
As a multi-step planning framework, we segment each task instance into several independent subtasks, treating each subtask as an action that can and can only be executed once.. Therefore, our action space decreases within an episode, after selecting action $a_i$ at the $i$-th step, $a_i$ will be removed from the action space $A_i$ to derive $A_{i+1}$. The selected subtask at each step is concatenated with the current state dictionary to serve as part of the prompt for the LLM.
The subtasks should meet the following criteria:
\begin{enumerate}
    \item[i.] Markov property: The execution of each subtask depends solely on the current State (the state contains the results from previous steps).
    \item[ii.] Completeness: Executing all subtasks in a certain order is equivalent to completing the original task, without omitting any requirements or text, and there is no overlap between subtasks.
    \item[iii.] Homogeneity: The type of subtasks should be consistent with the original task, and the structure of different subtasks within the same instance should be identical. This ensures the comparability among different actions when calculating Q-values, and allows for the use of the same prompt template.
\end{enumerate}

\paragraph*{Transition Function}
Since the state transition for a given action is completely determined by the LLM's output at this step, the generation probability of the LLM's response thus serves as the implicit state transition probability function $P$.

\paragraph*{Reward Function}
Based on the task type and the desired outcome, we design two types of rewards. 
For tasks where the final result is a concatenation of intermediate results, we use the stepwise reward. That is, at each step, the answer to the corresponding subtask is compared with the ground-truth to allocate a reward value. Moreover, if all the subtasks have equal weights (contributions to the final result), the discount factor $\gamma$ is set to 1. 
For tasks where the final result is derived from modifications of the intermediate results at each step, we employ an episode-level reward. There is no reward for the intermediate steps, instead, the reward is obtained based on the degree of match between the final result and the ground-truth. Consequently, the Q-values for intermediate steps is influenced by the discount factor and the final reward.

Some examples of state, action, and reward construction are shown in Table 1.

\subsection{Training Process of Actor Model}
\renewcommand{\algorithmicrequire}{\textbf{Input:}}
\renewcommand{\algorithmicensure}{\textbf{Parameter:}}
\begin{algorithm}[tb]
\caption{Training of the Actor model}
\label{algo:main}
\begin{algorithmic}[1] 
\REQUIRE Train sets $\mathcal{T}_1,...,\mathcal{T}_T$ with the same task type and language; \texttt{LLM}: the LLM embedded environment; \texttt{Rwd}: Reward Function; Initialized replay memory $\mathcal{D}$
\ENSURE $\theta$: network parameters; $\theta^\prime$: target-net parameters; $N_b$:training batch size; hyper-parameters $E,\epsilon,\gamma,k$
\FOR{epoch$=$\rm{1},...,$E$}
\STATE Sample instance $s$ from $T$ labeled train sets, with probability $1/T$ for each set. 
\STATE Split $s$ into several subtasks to obtain the action space $A_0$.
\STATE Get the task definition $d$, original text $C_0$, and corresponding requirements $R$.
\STATE Initialize the state $S_0=(d,C_0,I_0,R)$, where $I$ is the set of intermediate results, initialized to an empty set.
\FOR{t$=$\rm{0},...,$|A_0|-1$}
\STATE $p =$ \texttt{Random}$(0,1)$
\IF {$p\le$ \rm{1}-$\epsilon$}
\STATE $a_t = \arg\max_a Q(S_t,a;\theta)$
\ELSE
\STATE $a_t =$ \texttt{Random-sample}$(A_t)$.
\ENDIF
\STATE $A_{t+1}$=$A_t-\{a_t\}$
\STATE $I_{t+1}$=\texttt{LLM}$(S_t,a_t)$
\STATE Get ground truth $g_{t+1}$ from training sample $s$.
\STATE $r_{t+1}$=\texttt{Rwd}($I_{t+1},g_{t+1}$)
\STATE $S_{t+1}=(d,C_0,I_{t+1},R)$
\STATE Store $(S_t,a_t,r_{t+1},S_{t+1})$ in $\mathcal{D}$; randomly sample $N_b$ transitions $(S_j,a_j,r_j,S_{j+1})$ from $\mathcal{D}$, and do follows:
\IF{$S_{j+1}$ is terminal state}
\STATE $y_j=r_j$
\ELSE
\STATE $y_j=r_j+\gamma \max_{a}Q(S_{j+1},a;\theta^\prime)$
\ENDIF
\STATE $\mathcal{L}(\theta)=(y_j-Q(S_t,a_t;\theta))^2$
\STATE Update parameter $\theta$
\STATE Replace target-net parameters $\theta^\prime$ = $\theta$ every $k$ steps
\ENDFOR
\ENDFOR
\end{algorithmic}
\end{algorithm}

To model the linguistic features, such as grammatical structure and semantics for each instance, it is necessary to obtain sentence-level (sequence) representations. 
In each step, we flatten the state dictionary $S_t$ and concatenate each action in the action space $A_t$ with $S_t$ to form a natural language sequence: [\verb|<start>|, $a_{t_i}$, \verb|<sep>|, $S_t$, \verb|<end>|].

We can choose any pre-trained language model (PLM) as the foundation model to convert the tokens in the sequence into hidden vectors. For encoder PLMs such as BERT~\cite{devlin2018bert}, we select the first vector $h=h_0$ as the sequence representation, while for decoder PLMs, we select the last vector $h=h_n$.
A linear projection layer is added after the foundation model to constitute the Actor model, which learns the mapping from the sequence representation to the estimated Q-value, defined as:
\begin{equation}
	\hat{Q}(S_t,a_t) = \mathbf{W}h + \mathbf{b},
\end{equation}
where $ \mathbf{W}$ and $\mathbf{b}$ are trainable parameters.

We apply Deep Q-Learning to train the model and build a Deep Q-Network (DQN)~\cite{mnih2013playing}. We adopt the DDQN algorithm \cite{van2016deep}, utilize $\epsilon -$greedy exploration and the Experience Replay \cite{mnih2013playing} for RL training. 
According to Eq.~\ref{eq:Qlearn}, we design the iterative learning object as follows:
\begin{equation}
	Q(S,a) = r_{(S,a)} + \gamma \frac{1}{|\mathcal{S}_{(S,a)}|} \cdot \sum_{S^{\prime}\in \mathcal{S}_{(S,a)}}\max_{a^{\prime}\in A} Q(S^{\prime},a^{\prime}),
\end{equation}
where $r_{(S,a)}$ is the reward, $\mathcal{S}_{(S,a)}$ is the set of the successor states derived from the state $S$ with action $a$, and $\gamma$ is the discount factor.
The loss function is defined as the expected value of the mean squared Temporal Difference (TD) error:
\begin{equation}
    \mathcal{L} = \mathbb{E}(\hat{Q}(S,a)-Q(S,a))^2.
\end{equation}

The training process is illustrated in Algorithm \ref{algo:main}. In addition, for all datasets of the same language and task type, we only need to train a single Actor model, as their state structures are identical.

Once the Actor model is trained, it is capable of making plans based on the linguistic features of the task instance, selecting the subtask that maximizes the Q-value at each step:
\begin{equation}
\label{eq:infer_maxQ}
a^*_t=\arg\max_{a\in A_t} Q^*(S_t,a).
\end{equation}
The LLM executes the subtasks according to the optimal sequence $A^*$ and obtains the answer to the task instance. This interactive inferring process is illustrated in Algorithm~\ref{algo:infer}.

\renewcommand{\algorithmicensure}{\textbf{Output:}}
\begin{algorithm}[tb]
\caption{Interactive task solving process}
\label{algo:infer}
\begin{algorithmic}[1] 
\REQUIRE Task instance $s$; \texttt{LLM}: the LLM embedded environment; \texttt{Act}: the trained Actor model
\ENSURE Final result $I_F$, obtained by concatenating or modifying the intermediate results $I$
\STATE Split $s$ into several subtasks to obtain the action space $A$.
\STATE Get the task definition $d$, original text $C_0$, and corresponding requirements $R$; initialize the state $S=(d,C_0,I,R)$.
\WHILE{$A$ is not empty}
\STATE $a^*,Q^*$=None,-inf
\FOR{each $a$ in $A$}
\STATE $Q=\texttt{Act}(S,a)$
\IF{$Q>Q^*$}
\STATE $a^*,Q^*=a,Q$
\ENDIF
\ENDFOR
\STATE $A$=$A-\{a\}$; $I$=\texttt{LLM}$(S,a)$
\STATE Update state $S$
\ENDWHILE
\STATE Get final results $I_F$ from intermediate results set $I$.
\end{algorithmic}
\end{algorithm}

\section{RLAP Applications for Different NLP Tasks}
\label{sec:application}
In this section, we apply RLAP to three different NLP tasks, providing a detailed demonstration of their MDP modeling and the overall process for task completion. We conduct experiments to verify the effectiveness and robustness of the RLAP framework and offer insights for future research. The settings for states, actions, rewards and Actor models in different NLP tasks are shown in Table~\ref{tab:task_types}, and a brief introduction of the experimental datasets is provided in Appendix~\ref{sec:dataset}.
In addition, when applying RLAP for testing, we set the input of LLMs to be plain prompts without chain-of-thought (CoT) and in-context examples to highlight the performance improvement brought by our framework.
For each task, the test set is consistent across our method and baselines, and it has no overlap with the training data.


\begin{table*}[h]
\caption{Configuration of states, actions, rewards, and Actor models for different NLP tasks.}
\label{tab:task_types}
\centering
\small
\begin{tabularx}{\textwidth}{>{\raggedright\arraybackslash}X|>{\raggedright\arraybackslash}X|>{\raggedright\arraybackslash}X|>{\raggedright\arraybackslash}X|>{\raggedright\arraybackslash}X}
\hline
\textbf{Task Type} & \textbf{State} & \textbf{Subtasks} & \textbf{Actor model} & \textbf{Reward Function} \\ \hline
MRC (extractive QA or multiple-choice QA) & text of question + processed sentences (initialized as empty) + candidate sentences & Select the next sentence to process, and (in the final step) answer the question & gte-multilingual-base (305M) & Episode-level reward: r=1 if terminal state and the final answer is right; else 0 \\ \hline
IE (relational triple extraction) & original context + relation type + extracted elements (initialized as empty) & Select an element (s or o) and extract it from the original context & bert-base-uncased, bert-base-chinese (110M) & Stepwise \& Episode-level reward: r=1 if the extract result matches the ground-truth; else 0 \\ \hline
IE (event extraction) & original context + event type + roles schema + extracted role-argument pairs (initialized as empty) & Select a role according to the roles schema, and extract the corresponding argument from context & bert-base-uncased, bert-base-chinese (110M) & Stepwise \& Episode-level reward: r=1 if the extract role-argument pair matches the ground-truth; else 0 \\ \hline
STC (sentence to paragraph) & concatenated text (initialized as empty) + candidate sentences & Select a sentence and concatenate it to the existing text & qwen2.5 (7B) & Stepwise reward: r=1 if the completed text is a prefix of ground-truth, else 0 \\ \hline
STC (fill in the blanks) & incomplete context with several blanks + the sentence to be processed & Choose a suitable blank to fill in the sentence selected by the Actor model & gte-multilingual-base (305M) & Stepwise reward: r=1 if the sentence matches the blank, else 0 \\ \hline
\end{tabularx}
\end{table*}

\subsection{Machine Reading Comprehension}
\label{sec:MRC}
Machine Reading Comprehension (MRC) task is a challenging task and hot topic in NLP, where the goal is to answer the questions regarding a given context. In this paper, we consider two types of MRC tasks: the extractive QA requires extracting a continuous sequence of words from the text as the answer, while the multiple-choice QA requires selecting the most appropriate answer to the question from several options.

\paragraph*{Task setup}
We use SQuAD2.0 (English,~\cite{rajpurkar2018know}) and CMRC18 (Chinese,~\cite{cui-emnlp2019-cmrc2018}) for extractive QA task, RACE-H (English,~\cite{lai2017large}) and C3-mix (Chinese,~\cite{sun2019investigating}) for multiple-choice QA task.
These datasets contain multiple contexts from different domains, with each context corresponding to several questions. 
We select 2-3 questions for each context, treating a set of "question-answer-context" as a sample, and exclude contexts that are too short (less than 3 sentences). We randomly select 3,000-6,000 samples from each dataset for training the Actor model and 1,000-2,000 samples to construct the test set. We report the proportion of correct answers (i.e., the accuracy of answers) of the test set as the metric.

\begin{figure}[t]
\begin{center}
\includegraphics[width=0.9\linewidth]{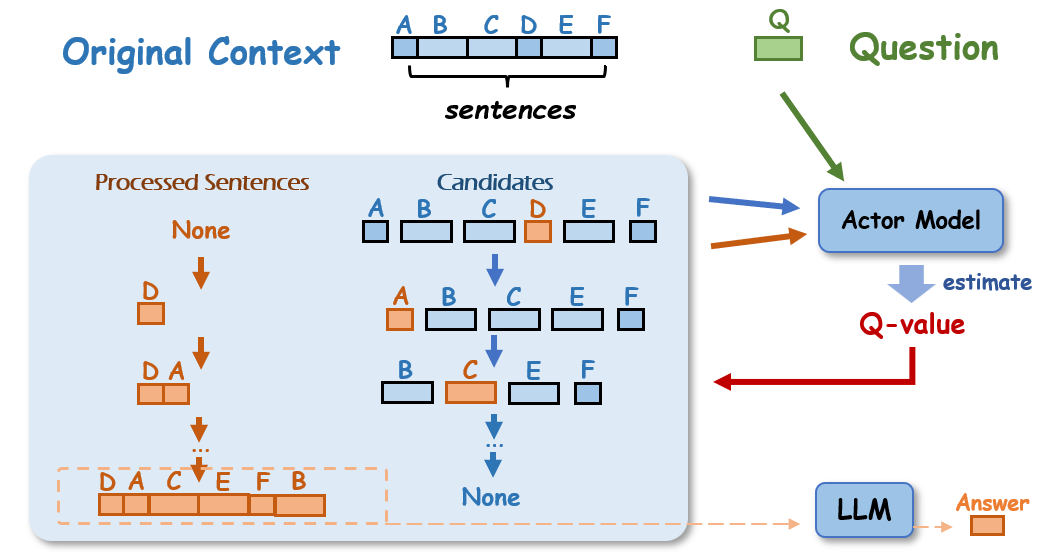} 
\caption{An illustration of RLAP in MRC tasks.}
\Description{None}
\label{fig:MRCdemo}
\end{center}
\end{figure}

\paragraph*{RLAP setup}
To frame the MRC tasks into our RLAP framework, we segment the context into sentences, with each sentence corresponding to a subtask. As shown in Fig.~\ref{fig:MRCdemo}, for a question to be answered, we initialize the "processed sentences" as empty and read one sentence at each step, marking it as "processed". Since not every sentence contains the information relevant to the question, the LLM answers the question only after all sentences have been read (i.e., at the final step).
In short, we seek an optimal reading order, which is a permutation of the sentences in the original context, and then answer the question after reading all sentences in this order. 
The motivation is that when humans answer questions, they tend to prioritize reading the most relevant sentences in the context based on the content of the question (for example, opinions often appear at the beginning or end of a paragraph). Therefore, the reading order needs to be adjusted according to the linguistic features of the question and the context, as shown in Fig.~\ref{fig:intro_1}.
We add a linear layer of 768-dimensional to 1-dimensional after the last layer of \textit{gte-multilingual-base}~\cite{zhang2024mgte} to constitute the Actor model, with all layer parameters being trainable. During training, we compare the LLM's response at the final step with the standard answer to obtain an episode-level reward.

\paragraph*{Baselines}
We employ the standard input-output (IO) prompt with 3 in-context examples and directly input the context and question to the LLM to obtain the answer.
Additionally, we apply ToT-BFS~\cite{yao2024tree} with b=3, which means exploring three different sentences (branches) at each step. The LLM autonomously evaluates the responses after reading each branch and selects the most suitable one. Then, the sentence corresponding to this branch is removed from the candidate set, and the search continues to the next step. When the candidate set is empty, we get the final answer.

\paragraph*{Experimental results}
We use Qwen2.5-14B~\cite{qwen2.5} as the LLM embedded in the environment. As shown in Table~\ref{tab:result_MRC}, RLAP significantly outperforms both standard IO and ToT-BFS on all datasets, with the highest accuracy improvement reaching 2.45\% over the baselines.
In addition, we construct a test set for complex scenarios using contexts with six or more sentences. In complex settings, RLAP's performance remains stable and superior to the baselines, with even more significant improvements (up to 3.92\%). The experimental results validate the effectiveness of our method on MRC tasks and its stability in complex situations.

\begin{table}[t]
	\caption{The accuracy of answers on MRC tasks. “general case” refers to the results on all test samples, while “complex case” refers to the results on test samples where the context contains 6 or more sentences.}
	\label{tab:result_MRC}%
	\centering
	\resizebox{1.00\columnwidth}{!}{
		\begin{tabular}{p{1.2cm}p{2.5cm}cccc}
			\toprule
		Cases & \diagbox{Methods}{Datasets} & SQuAD2.0  & CMRC18 & C3-mix & RACE-H \\
			\midrule
			\multirow{4}{*}{\text{ }\text{ }\rotatebox{90}{general}} & IO prompt & 89.39 & 64.88 & 77.85 & 91.70 \\
            & TOT-BFS (b=3) & 89.52 & 64.21 & 77.97 & 89.20  \\   
            & RLAP (ours) & \textbf{91.71} & \textbf{64.92} & \textbf{79.64} & \textbf{91.90}  \\
    & \textit{Improvement} & +2.45\% & +0.06\% & +2.14\% &  +0.22\%  \\

   \midrule
			\multirow{4}{*}{\text{ }\text{ }\rotatebox{90}{complex}} & IO prompt & 88.76 & 63.54 & 77.92 & 85.05 \\
            & TOT-BFS (b=3) & 87.62 & 63.54 & 77.37 & 85.93  \\   
            & RLAP (ours) & \textbf{91.19} & \textbf{64.77} & \textbf{78.50} & \textbf{89.30}  \\
    & \textit{Improvement} & +2.74\% & +1.94\% & +0.74\% &  +3.92\%  \\
			\bottomrule
		\end{tabular}%
	}
\end{table}%

\subsection{Information Extraction}
Information Extraction (IE) task plays an important role in knowledge acquisition, and we focus on relation extraction (RE) and event extraction (EE) tasks in this section. 
The RE task aims to extract triples (subject, predicate, object) from the context, while the EE task involves extracting the arguments corresponding to each role of a given event type from the context.

\paragraph*{Task setup}
We use NYT10 (English,~\cite{riedel2010modeling}) , HacRED (Chinese,~\cite{hacred}) and SKE21 (Chinese,~\cite{xie-etal-2021-revisiting}) for RE task, ACE05\footnote{\url{https://catalog.ldc.upenn.edu/LDC2006T06}} (English) and DuEE (Chinese,~\cite{li2020duee}) for EE task.
These datasets contain a variety of prespecified relation or event types, and each context may include multiple relations/events. We treat a set of "relation/event type-context-\{slots:arguments\}" as a sample, where \textit{slots} refer to the entities to be extracted (for RE, slots are the subject and object; for EE, slots are the roles of the event). 
We randomly select 10,000 samples from each dataset to train the Actor model and choose 1,000-1,500 samples to construct the test set. 
We report the F1-score based on the \textbf{exact match} for evaluation, considering a slot extraction as a true positive only when the result matches the ground-truth exactly.

\begin{table*}[htb]
	\caption{The F1 score of extraction results on IE tasks. In the complex case, we select contexts with more than 10 triples in Hacred test set, contexts with more than 3 triples in NYT10 test set, and samples with more than 4 slots in DuEE test set.}
	\label{tab:IE_mainresult}%
	\centering
	\resizebox{1.75\columnwidth}{!}{
		\begin{tabular}{p{1cm}p{2.5cm}cccccccc}
			\toprule
		\multirow{2}{*}{LLMs} & \multirow{2}{*}{\diagbox{Methods}{Datasets}} & \multicolumn{5}{c}{General Case} & \multicolumn{3}{c}{Complex Case} \\
        \cmidrule(r){3-7} \cmidrule(r){8-10}
        & & HacRED & NYT10 & SKE21 & DuEE & ACE05 &  HacRED  & NYT10 & DuEE \\
			\midrule
			\multirow{6}{*}{\text{ }\text{ }\rotatebox{90}{Qwen2.5-14B}} & ChatIE & 49.36 & 39.35 & 46.79 & 65.73 & 31.57 & 43.51 & 32.19 & 59.37 \\
            & RL4IE & 58.83 & 52.97 & 75.83 & 69.80 & 68.93 & 55.61 & 50.18 & 66.83 \\   
            & RLAP-RL (ours) & \textbf{77.27} & \textbf{54.81} & \textbf{83.64} & \textbf{77.33} & \textbf{75.83} & \textbf{75.12} & \textbf{52.97} & \textbf{74.91}\\
    & \textit{Improvement} & +31.3\% & +3.5\% & +10.3\% & +10.8\% & +10.0\% & +35.1\% & +5.6\% & +12.1\% \\
    \cmidrule{2-10}
            & RLAP-random & 73.33 & 53.31 & 78.87 & 74.16 & 70.32 & 71.49 & 49.06 & 72.20 \\
            & RLAP-sequence & 70.05 & 46.54 & 73.03 & 73.38 & 71.40 & 66.40 & 41.30 & 71.44 \\

   \midrule
			\multirow{6}{*}{\text{ }\text{ }\rotatebox{90}{Mistral-7B}} & ChatIE & 11.73 & 18.02 & 13.80 & 38.10 & 49.27 & 9.03 & 17.92 & 30.94\\
            & RL4IE & 16.41 & 62.27 & 29.43 & 59.03 & 54.92 & 13.53 & 59.91 & 58.18 \\   
            & RLAP-RL (ours) & \textbf{23.47} & \textbf{63.17} & \textbf{36.13} & \textbf{63.33} & \textbf{55.30} & \textbf{21.18} & \textbf{62.09} & \textbf{61.32}\\
    & \textit{Improvement} & +43.0\% & +1.4\% & +22.8\% & +7.3\% & +0.7\% & +56.5\% & +3.6\% & +5.4\% \\
    \cmidrule{2-10}
            & RLAP-random & 15.44 & 61.53 & 29.61 & 60.99 & 45.88 & 9.16 & 59.38 & 58.89\\
            & RLAP-sequence & 18.03 & 59.81 & 20.57 & 56.26 & 54.69 & 10.99 & 57.99 & 53.96\\
			\bottomrule
		\end{tabular}%
	}

\end{table*}%

\begin{figure}[!ht]
\begin{center}
\includegraphics[width=1.0\linewidth]{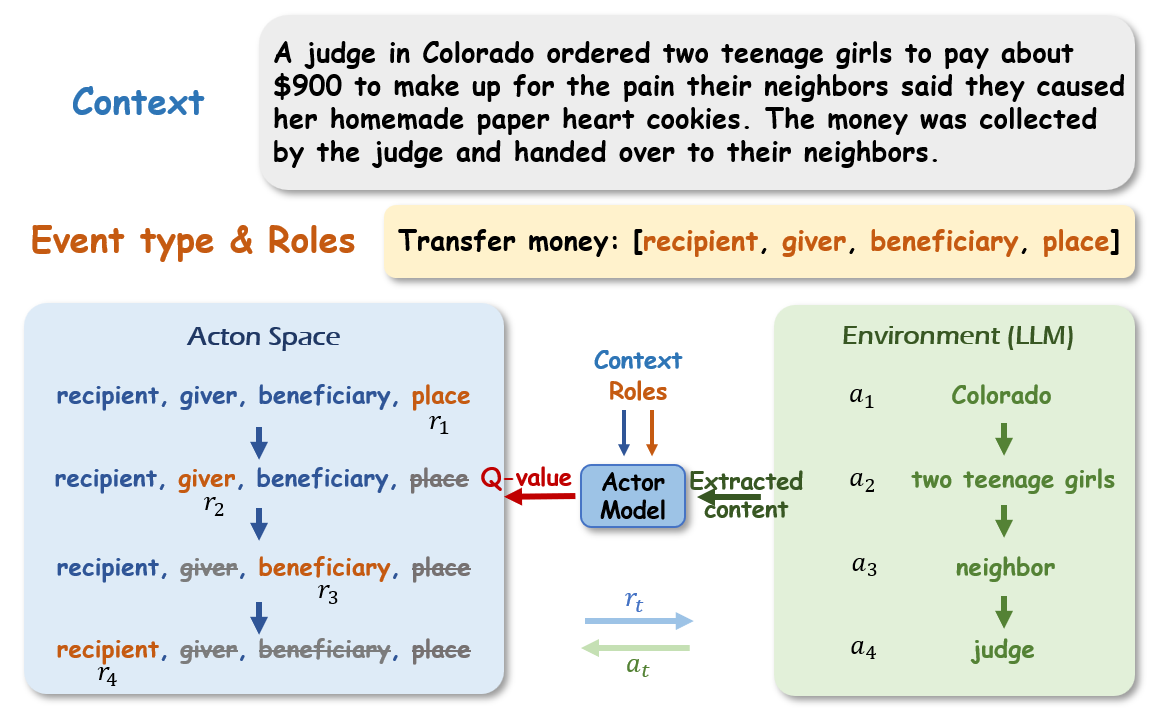} 
\caption{An example of RLAP in IE tasks.}
\Description{None}
\label{fig:IEdemo}
\end{center}
\end{figure}

\paragraph*{RLAP setup}
We treat each slot as an action, and the subtask is to extract the argument corresponding to a slot. As shown in Fig.~\ref{fig:IEdemo}, at each step, the Actor model estimates the Q-value according to the context, the role schema, and the extracted results, decides the slot to be extracted next and removes it from the action space. 
Meanwhile, the LLM extracts the corresponding argument and adds the "slot-argument" pair to the extracted results. The extraction is completed when the action space is empty.
Previous work on IE tasks has attempted to automatically select the slot to extract~\cite{ding2024adaptive}, but they use text similarity to evaluate the extraction results, introduce an additional LLM as the reward model, and only consider stepwise rewards. It is inaccurate, time-consuming, and derives the reward from the LLM's classification result, which is prone to errors. In addition, it lacks an overall assessment of the results for each episode.
Therefore, during training, we directly compare the extraction results at each step with the ground truth to obtain a stepwise reward. Additionally, we compare the overall extraction results of the sample to get the episode-level reward, which is 1 if all slots are correctly extracted and 0 otherwise. 
We constructed the Actor model by adding a mapping layer after the last layer of the BERT~\cite{devlin2018bert} model (\textit{bert-base-uncased} for English and \textit{bert-base-chinese} for Chinese).

\paragraph*{Baselines}
We apply the ChatIE ~\cite{wei2023zero} multi-turn instruction template to open-source LLMs, with 5 in-context samples. We also use the previous method that employs an LLM as a reward model and only considers step-wise rewards (denoted RL4IE in Table~\ref{tab:IE_mainresult})~\cite{ding2024adaptive} as a baseline.

\paragraph*{Experimental results}
We employ two open-source LLMs, Qwen2.5-14B~\cite{qwen2.5} and Mistral-7B-instruct-v0.3\footnote{\url{https://huggingface.co/mistralai/Mistral-7B-Instruct-v0.3}}, for the experiment. 
As shown in Table~\ref{tab:IE_mainresult}, RLAP has better extraction capabilities than baselines on different datasets, achieving the highest performance improvement of 43.0\%. 
Additionally, we construct test sets for complex scenarios by controlling the number of triples per context for the IE task (more than 10 triples for HacRED and more than 3 triples for NYT) and the number of slots per sample for the EE task (more than 4 slots for DuEE). In complex settings, the performance of RLAP remains better than baselines, with the highest improvement reaching 56.5\%.

\paragraph*{Ablation study}
To examine the role of the Actor model in the RLAP framework, we experiment with two ablation settings, namely \textit{RLAP-random} and \textit{RLAP-sequence} as shown in Table~\ref{tab:IE_mainresult}. 
The "random" setting involves randomly selecting the next slot to be extracted at each step, while the "sequence" setting uses the same pre-specified order of slots for extraction across all instances of the same relation/event type. Neither of these settings employs the Actor model or considers the linguistic features of the instances.
The experimental result shows that the extraction results under the ablation settings are inferior to those of the standard RLAP framework, either in general or in complex cases. It confirms the effectiveness and necessity of incorporating the Actor model for adaptive order planning.

\subsection{Sentence-level Text Completion}
Sentence-level Text Completion (STC) task is to generate a semantically and grammatically correct context based on candidate sentences. 
In this section, we consider two STC tasks: sentence-to-paragraph (S2P) and sentence-level filling-in-blanks (SFB, also known as Sentence Cloze-Style MRC). 
The S2P task involves concatenating the given candidate sentences into a fluent paragraph. The SFB task provides a context with several blanks, and we need to select the most appropriate sentence from the candidates for each blank to finally form a complete context.

\paragraph*{Task setup}
For the S2P task, we select contexts that contain more than 3 and less than 8 sentences from the HacRED (Chinese,~\cite{hacred}) and SQuAD2.0 (English,~\cite{rajpurkar2018know}) datasets, and regard these contexts as the ground-truth. 
Each context is split into sentences and shuffled to form a set of candidates as a single sample. We randomly select 1,000-3,000 samples from each dataset for training the Actor model and select 500 samples as the test set. 
We use context-level accuracy (CAC) and sentence-order accuracy (SOC) as evaluation metrics, defined as follows:
\begin{equation*}
\small
\text{CAC}=\frac{\#\text{ correct final results}}{\#\text{ contexts}}; \text{SOC}=\frac{1}{\#\text{ contexts}}\sum\frac{\#\text{ correct pairs}}{C_n^2},
\end{equation*}
where $n$ denotes the number of candidate sentences in one context, and "correct pairs" refer to pairs of sentences whose order matches the order in the ground-truth.
For the SFB task, we select contexts with more than 5 blanks from the CMRC19 (Chinese,~\cite{cui-etal-2020-cmrc2019}) train and trial datasets. Each set of "incomplete context-candidate sentences" is treated as a single sample. We randomly select 2,500 samples for training the Actor model and 500 samples for the test set. We use Blank-level Accuracy (BAC) and CAC to evaluate the results, and the BAC is defined as follows:
\begin{equation*}
\small
\text{BAC}=\frac{\#\text{ correct filled blanks}}{\#\text{ total blanks}}.
\end{equation*}

\begin{figure}[t]
\begin{center}
\includegraphics[width=\linewidth]{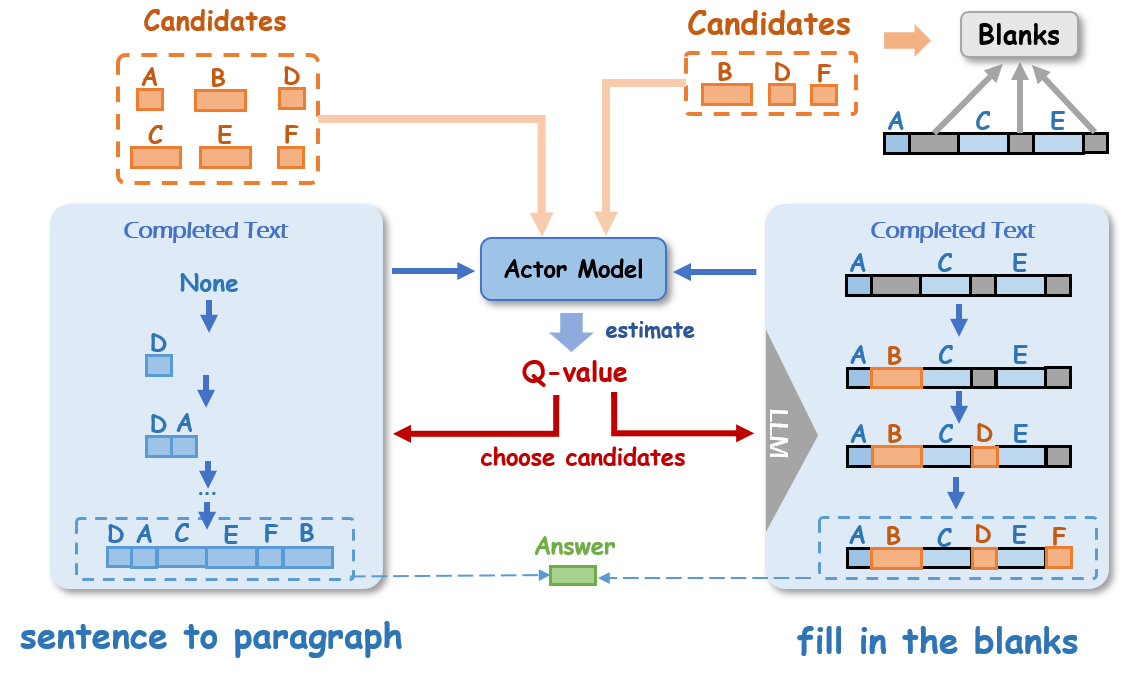} 
\caption{An illustration of RLAP in STC tasks.}
\Description{None}
\label{fig:STCdemo}
\end{center}
\end{figure}

\paragraph*{RLAP setup}
For STC tasks, we regard candidate sentences as the action space. At each step, the Actor model selects a sentence from it according to Eq.~\ref{eq:infer_maxQ}, and the subtask is to concatenate this sentence to the end of the completed text or fill it into the appropriate blank, as shown in Fig.~\ref{fig:STCdemo}.
For the S2P task, since the subtask is a trivial text concatenation, we do not need to embed LLMs in the environment. Additionally, to demonstrate the universality of RLAP, we select Qwen2.5-7B, a lightweight decoder LLM, as the foundation model, and add a 3584-dimensional to 1-dimensional linear mapping layer after its last layer to construct the Actor model.
To reduce training complexity, we only train the parameters of the additional linear mapping layer, keeping the parameters of the foundation model frozen. 
We adopt a stepwise reward, which is 1 at each step if the completed text is a prefix of the ground-truth (i.e., the sentences in the completed text are in the correct order) and 0 otherwise.
For the SFB task, we construct the Actor model based on \textit{gte-multilingual-base} as in Section~\ref{sec:MRC}. At each step, the LLM fills the selected sentence into the appropriate blank to replace the \verb|[blank]| symbol and returns the updated context. The task is completed when all the blanks are filled in. During training, we obtain stepwise rewards based on whether the selected sentence matches the blank in each step.

\paragraph*{Baselines}
For the S2P task, since the Actor model adds a small number of parameters (the mapping layer) on top of Qwen2.5-7B, we choose Qwen2.5-14B with CoT prompt (without examples and with 3 in-context examples) as baselines. We also randomly select a sentence for concatenation at each step as a baseline (denoted "random action" in Table~\ref{tab:result_S2P}), and its CAC is expected to be small and SOC is expected to be 50\%.
For the SFB task, we use a CoT prompt with 3 in-context examples to guide LLM to fill in the blanks step by step. The sentence in each step is randomly selected (denoted CoT-random) or sequentially selected (denoted CoT-sequence).

\begin{table}[t]
	\caption{The CAC and SOC of final results on S2P task.}
	\label{tab:result_S2P}%
	\centering
	\resizebox{0.9\columnwidth}{!}{
		\begin{tabular}{p{2.5cm}cccc}
			\toprule
		\multirow{2}{*}{\diagbox{Methods}{Datasets}} & \multicolumn{2}{c}{HacRED} & \multicolumn{2}{c}{SQuAD2.0} \\
        \cmidrule{2-5}
        & CAC(\%) & SOC(\%) & CAC(\%) & SOC(\%) \\
			\midrule
            CoT (0 shot) & 16.89 & 60.45 & 17.40 & 56.54 \\
            CoT (3 shot) & 17.56 & 61.33 & 19.60 & 61.08 \\ 
            random action & 6.52 & 48.90 & 8.20 & 51.16 \\ 
            RLAP (ours) & \textbf{23.58} & \textbf{76.42} & \textbf{22.37} & \textbf{71.37} \\ 
            \textit{Improvement} & +34.3\% & +24.6\% & +14.1\% &  +16.8\%  \\
			\bottomrule
		\end{tabular}%
	}
\end{table}%


\begin{figure}[t]
\vspace{0.2cm}
	\centering
	\begin{minipage}[t]{0.235\textwidth}
        \centering
        \includegraphics[width=\linewidth]{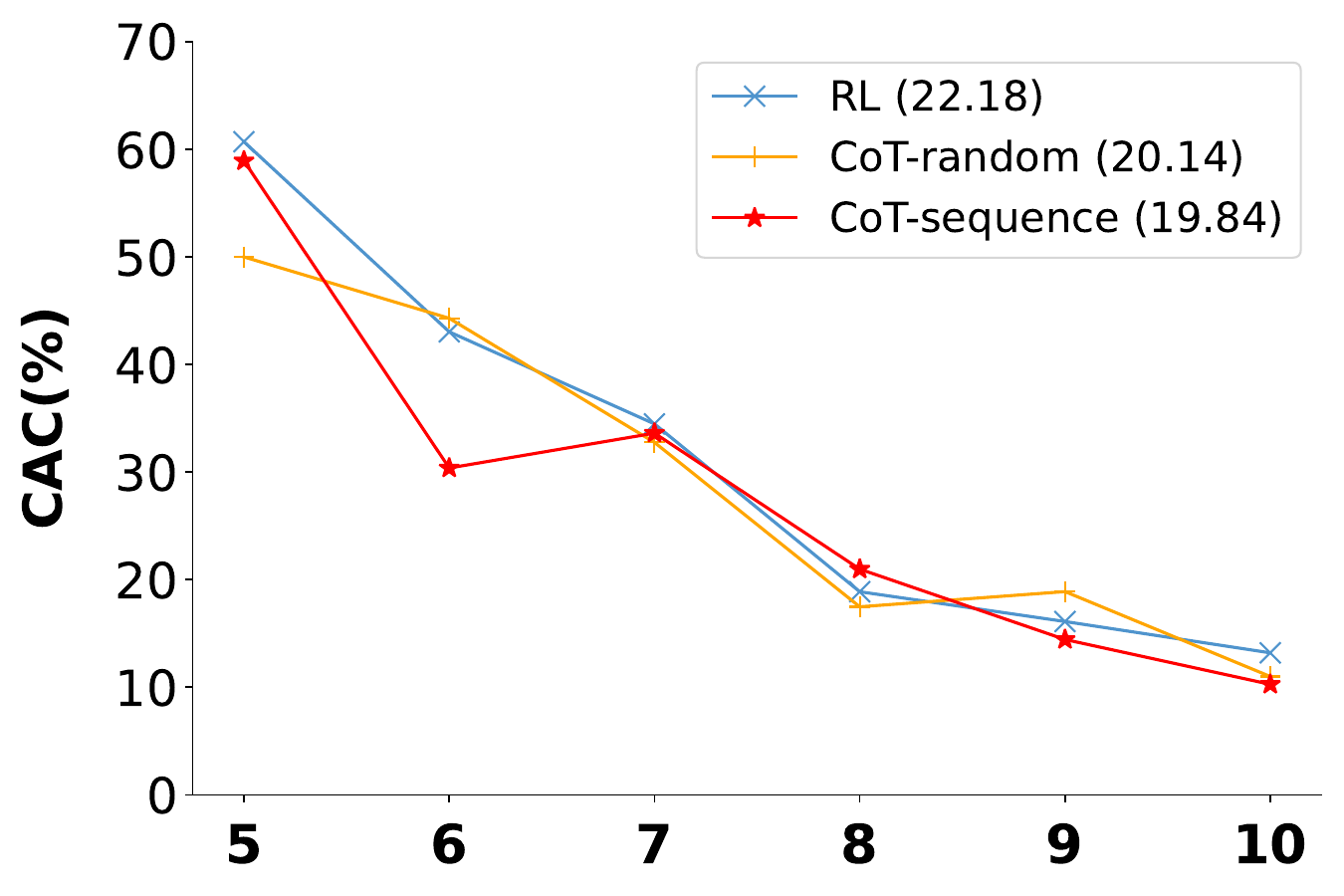}
    \end{minipage}
    \begin{minipage}[t]{0.235\textwidth}
        \centering
        \includegraphics[width=\linewidth]{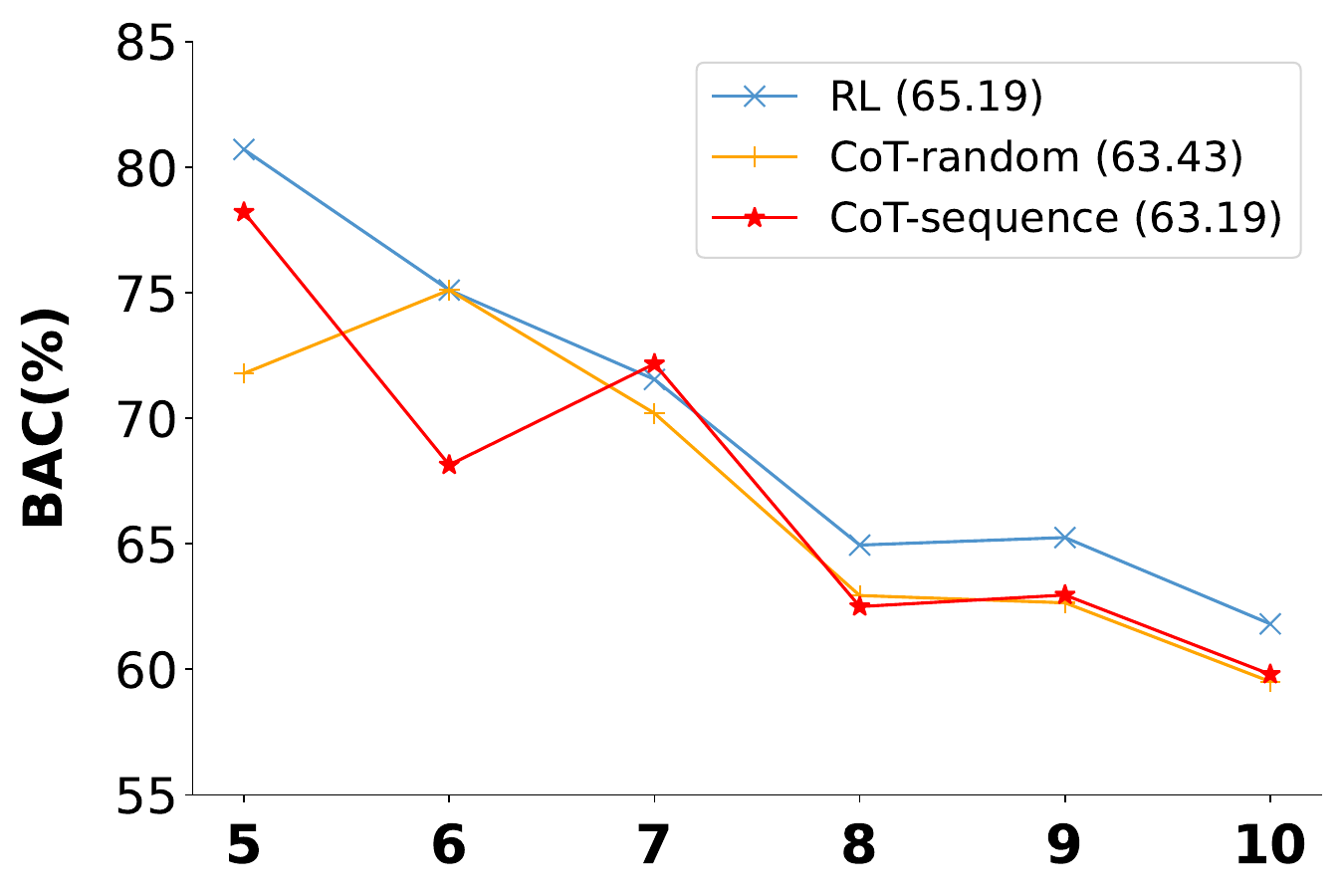}
    \end{minipage}
	\caption{The X-axis represents the number of blanks per sample, and the numbers in the legend indicate the metric values on the entire test set.}
	\label{fig:STC_complex}
\end{figure}

\paragraph*{Experimental results}
For the S2P task, Table~\ref{tab:result_S2P} shows that the results of random-action are as expected, and RLAP outperforms the baselines on both Chinese and English datasets, with significant improvements in both CAC and SOC.
For the SFB task, we employ Qwen2.5-14B as the LLM embedded in the environment. As shown in Fig.~\ref{fig:STC_complex}, we report the CAC and BAC calculated on the entire test set, with RLAP outperforming the baselines by approximately 2\%. In addition, we check the trend of metrics with the number of blanks. 
By filtering the test set based on the number of blanks in the context and calculating the metrics separately, we observe that both CAC and BAC decrease as the number of blanks increases, and in most cases, RLAP performs better than the baselines.

\section{Conclusion}
In this paper, we investigate the impact of the linguistic features of task instances on multi-step NLP task solving and propose a Reinforcement Learning Enhanced Adaptive Planning Framework for LLM (RLAP).
We first generally introduced the components of RLAP, including the construction of states, the selection of subtasks, the design of rewards, and the structure and training of the Actor model, which maps linguistic features to Q-values to select the optimal action at each step. 
Then, we select three different types of NLP tasks to demonstrate the practical application of RLAP in detail and set up baselines for comparative experiments. The experimental results validate the effectiveness, universality, and stability of RLAP.
In future work, we will further explore the application of RLAP during the pre-training phase and in multimodal tasks.

\newpage
\bibliographystyle{ACM-Reference-Format}
\bibliography{sample-base}

\newpage
\appendix

\section{Dataset Introduction}
\label{sec:dataset}
\paragraph*{SQuAD2.0}
Stanford Question Answering Dataset (SQuAD) is a reading comprehension dataset, consisting of questions posed by crowdworkers on a set of Wikipedia articles, where the answer to every question is a segment of text, or span, from the corresponding reading passage, or the question might be unanswerable. SQuAD2.0~\cite{rajpurkar2018know} combines the 100,000 questions in SQuAD1.1 with over 50,000 unanswerable questions written adversarially by crowdworkers to look similar to answerable ones.

\paragraph*{CMRC18}
CMRC2018~\cite{cui-emnlp2019-cmrc2018} is a Span-Extraction dataset for Chinese machine reading comprehension to add language diversities in this area. The dataset is composed by near 20,000 real questions annotated on Wikipedia paragraphs by human experts, and the questions need comprehensive understanding and multi-sentence inference throughout the context. The answer is a continuous span in the context.

\paragraph*{CMRC19}
CMRC2019~\cite{cui-etal-2020-cmrc2019} is a Chinese sentence-level cloze-style dataset. It contains over 100K blanks within over 10K passages, which was originated from Chinese narrative stories. Each sample provides a narrative passage and several sentences extracted from it. The model is required to precisely fill the candidate sentences back into the blanks in the original passage to form a complete article.

\paragraph*{RACE-H}
RACE~\cite{lai2017large} is a large-scale reading comprehension dataset collected from English examinations designed for 12–15 year-old middle school students (RACE-M), and 15–18 year-old high school students (RACE-H) in China. RACE consists of near 28,000 passages and near 100,000 questions generated by human experts (English instructors), and covers a variety of topics which are carefully designed for evaluating the students’ ability in understanding and reasoning.

\paragraph*{C3-mix}
C3~\cite{sun2019investigating} is short for the multiple-Choice Chinese machine reading Comprehension dataset, it contains 13,369 documents from dialogues (C3-dialog) or more formally written mixed-genre texts (C3-mix) and their associated 19,577 free-form multiple-choice questions collected from Chinese-as-a-second-language examinations.

\paragraph*{HacRED} 
HacRED~\cite{hacred} is a novel challenging extraction dataset. It analyzes the performance gap between popular datasets and practical applications, and carefully selects and designs more hard cases. HacRED consists of 65,225 relational facts annotated from 9,231 wiki documents with sufficient and diverse hard cases, which poses a very high challenge to many current complex extraction methods.

\paragraph*{NYT10}
NYT is based on the articles in New York Times. There are many derived datasets with better labeling. NYT10~\cite{riedel2010modeling} labels the complete entities.

\paragraph*{SKE21}
SKE19\footnote{\url{http://ai.baidu.com/broad/download?dataset=sked}} is published by Baidu, and is currently the largest dataset available for complex relational triple extraction. Since its testing set is unpublished, and there are some errors in the validation set, a version named SKE21 is published by \citet{xie-etal-2021-revisiting}. The testing set of SKE21 is carefully manually relabeled and contains 1,150 sentences and 2,765 annotated triples.

\paragraph*{DuEE}
DuEE \cite{li2020duee} is a Chinese document-level event extraction dataset. It contains 11, 224 documents categorized into 65 event types, along with 41, 520 event arguments mapped to 121 argument roles, which is the largest Chinese EE dataset.

\paragraph*{ACE05}
ACE2005\footnote{\url{https://catalog.ldc.upenn.edu/LDC2006T06}} Multilingual Training Corpus was developed by the Linguistic Data Consortium (LDC) and contains approximately 1,800 files of mixed genre text in English, Arabic, and Chinese annotated for entities, relations, and events. In this paper we use the English events corpus, it provides event annotations in document and sentence levels from a variety of domains such as newswires and online forums.





\section{Experiment Details}
Our experiments are conducted on two A800 GPUs (mainly used for deploying LLMs). All LLMs embedded in the environment have not undergone any form of fine-tuning, and their timeout is set to 6 seconds. 

The Actor models are implemented using the PyTorch framework. For the RL training of the Actor model, we set the buffer size to 5000 and the target network update step to 20. We train 10 epochs for all train set. Additionally, the exploration parameter $\epsilon$ is initialized at 0.9 and the discount factor $\gamma$ is set to 0.5. The exploration rate $\epsilon$ becomes 0.95 times its own in every 100 steps until it reaches 0.02.

Our code, detailed configuration, and prompt templates can be found at \url{https://anonymous.4open.science/r/RLAP-anonymized-2545}.

\end{document}